\pdfoutput=1

\documentclass[11pt]{article}

\usepackage[]{ACL2023}

\usepackage{times}
\usepackage{latexsym}

\usepackage[T1]{fontenc}

\usepackage[utf8]{inputenc}

\usepackage{microtype}
\usepackage{booktabs}
\usepackage{microtype}
\usepackage{amsmath}
\usepackage{amsmath,amssymb, amsthm, amstext}
\usepackage{mathrsfs}
\usepackage{extarrows}
\usepackage{multirow}
\usepackage{graphicx}
\usepackage{booktabs}
\usepackage{caption}
\usepackage{subcaption}
\usepackage{enumitem}
\usepackage{arydshln}
\usepackage{xcolor}
\usepackage{svg}
\usepackage{tikz}
\usepackage{tabularx}
\usepackage{wasysym}
\usepackage{latexsym}

\title{\texttt{Koala}: An Index for Quantifying Overlaps with  Pre-training Corpora}

\author{Thuy-Trang Vu$^{{\ddagger} }$\ \ \ \ \ \ \ \ \ Xuanli He$^{{\dagger}}$\ \ \ \ \ \ \ \ \ Gholamreza Haffari$^{{\ddagger} }$\ \ \ \ \ \ \ \ \ Ehsan Shareghi$^{{\ddagger} }$\Thanks{~~Corresponding author}\\ \ \ \
$^{{\ddagger} }$Department of Data Science \& AI, Monash University\\
$^{\dagger}$Department of Computer Science, University College London\\
\texttt{\small{\{trang.vu1, gholamreza.haffari, ehsan.shareghi\}@monash.edu}} \ \ \ \texttt{\small{h.xuanli@ucl.ac.uk}}}

\begin{document}
\maketitle
\begin{abstract}
In very recent years more attention has been placed on probing the role of pre-training data in Large Language Models (LLMs) downstream behaviour. Despite the importance, there is no public tool that supports such analysis of pre-training corpora at large scale. To help research in this space, we launch \texttt{Koala}, a searchable index over large pre-training corpora using compressed suffix arrays with highly efficient compression rate and search support. In its first release we index the public proportion of OPT 175B pre-training data.
\texttt{Koala} provides a framework to do forensic analysis on the current and future benchmarks as well as to assess the degree of memorization in the output from the LLMs.  
\texttt{Koala} is available for public use at \url{https://koala-index.erc.monash.edu/}
\end{abstract}

\section{Introduction}
Large Language Models (LLMs) have achieved state-of-the-art results in NLP and on many benchmarks  have reached the performance ceiling~\cite{chowdhery2022palm}. This evergrowing success has been facilitated by the algorithmic and computational progress in scaling up model sizes~\cite{wei2022emergent,chowdhery2022palm,zhang2022opt,brown2020language}, integrating human feedback~\cite{ouyang2022training}, adopting modes of instructional inference at both zero- or few-shot settings~\cite{chen2022program, kojima2022large,wei2022chain,nye2021show}, as well as the ability of feeding them massive volumes of free text during pre-training.  

Recent works exhibit various cases which highlight the sensitivity of downstream behaviour of LLMs (and their smaller variants)  to the frequency of observed overlap between pre-training corpora and test set~\cite{carlini2022quantifying,tanzer2022memorisation,razeghi2022impact,magar2022data,lewis2020question}. In the generative setting, several issues such as hallucination~\cite{dziri-etal-2022-origin}, undesired biases~\cite{kirk2021bias}, or toxicity~\cite{gehman-etal-2020-realtoxicityprompts} have been attributed partly or fully to the characteristics of 
the pre-training data, while a parallel line of works have emphasised on the positive role of filtering the pre-training data for safety and factual grounding~\cite{thoppilan2022lamda}.

The above observations are not a comprehensive list but echo \emph{the undeniable role of pre-training data in how these models would function in practice}. Understanding the limitations imposed by pre-training data would  also lead to more informed algorithmic and computational innovations~\cite{DBLP:journals/corr/abs-2208-11981}. However, these forensic studies are done either at a small scale or by using surrogate sources such as web search hit counts. To the best of our knowledge, there is no existing work that offers a tool or service for supporting deeper analyses in this space at large scale. 

To help research in this direction, we launch the \texttt{Koala} project, a service backed by lossless compressed suffix arrays (CSA)~\cite{nv-csurv07},
 with efficient compression rate and query support. \texttt{Koala}  contains a searchable index over the public portion of the pre-training data\footnote{We plan to extend our coverage of pre-training corpora.} behind the OPT 175B \cite{zhang2022opt} model, after deduplication. The constructed index is intended to provide overlap count statistics for text query (with $\infty$-gram length) files provided by users. We foresee several areas of impact for \texttt{Koala}; (i) as a tool to measure data leakage between existing benchmarks and pre-training corpora of LLMs, (ii) and evaluate the degree of memorisation or creativity in generative models' output, (iii) and to support designing harder benchmarks by reducing the  overlap with pre-training corpora.     

We present a brief overview of the \texttt{Koala} pipeline for pre-processing and constructing the index. We also provide examples of the types of analyses that could be done via \texttt{Koala} by looking at a few commonly used test benchmarks.  

\section{Pre-processing and Corpora Coverage}
\subsection{Pre-processing Steps}\label{sec:deduplication}
Our pre-processing pipeline includes three main steps: cleaning, deduplication and tokenization. The cleaning step varies according to the pre-trained corpus and is described in Section~\ref{sec:corpus} where we provide the coverage of \texttt{Koala}. In this section, we describe the deduplication and tokenization steps which are shared across all pretrained corpora.

We use MinHashLSH~\citep[Chapter~3]{Rajaraman&Ullman}, a widely-adopted duplicate detection method for large-scale dataset, in the deduplication step. Documents are first converted into a set of unigram tokens (shingling) and then  hashed into a short signature, namely minhash, such that the similarity among documents is preserved. MinHash is a hashing algorithm based on permutation to generate random hashes to approximate the Jaccard similarity~\citep{broder1997resemblance,908981}. We generate the minhashes with 100 permutations. 
Finally, the locality-sensitive hashes (LSH) of the minhash values are calculated to detect the duplicated candidate pairs. We follow \citet{zhang2022opt} to remove those having Jaccard similarity scores above 0.95 threshold. Our deduplication implementation is based on the datasketch library\footnote{\url{https://github.com/ekzhu/datasketch}}. To scale the deduplication process to the large corpus, we first perform deduplication in a small batch and gradually merge the deduplicated batches.

The deduplicated corpus is then tokenized with moses~\citep{koehn-etal-2007-moses} to normalize punctuation and remove non-printing characters. We do not apply any  casefolding.

\subsection{Corpora Coverage}\label{sec:corpus}
The latest version of \texttt{koala} at the time of writing this manuscript covers the following corpora:\footnote{We plan to index more public pre-training corpora as they become available.}
\begin{description}[leftmargin=2.5mm,parsep=0pt,partopsep=0pt]
    \item [BookCorpus]\cite{Zhu_2015_ICCV} obtained from Hugging Face.\footnote{\url{https://huggingface.co/datasets/bookcorpus}}
    \item [CCNewsv2] extracted English news published between 2016 and 09/2021 from CommonCrawl~\citep{nagel2016cc} using news-please~\citep{Hamborg2017}.
    \item [ThePile]\citep{DBLP:journals/corr/abs-2101-00027} includes a subset of The Pile: Pile-CC, USPTO Backgrounds, Guthenberg~\cite{DBLP:conf/iclr/RaePJHL20}, OpenWebTexts \cite{Gokaslan:2019}, OpenSubtitles~\cite{DBLP:conf/lrec/Tiedemann16}, Wikipedia (en), DM Mathematics~\cite{DBLP:conf/iclr/SaxtonGHK19}, HackerNews.
    \item [Pushshift Reddit]\footnote{\url{https://files.pushshift.io/reddit}} We used langdetect\footnote{\url{https://github.com/fedelopez77/langdetect}} to detect and extract the English comments and submission posted from 2005 to 2019. We followed pre-processing procedure in \citep{roller-etal-2021-recipes} to remove the post from known non-English subreddits and bot\footnote{\url{https://github.com/eliassjogreen/Reddit-Bot-List}}, comments longer than 2048 characters or containing URL, or at depth larger than 7 in a thread.
\end{description}
For readers' reference, the above collection covers the pre-training corpora of OPT~\cite{zhang2022opt} with the exception of CC-Stories~\cite{Trieu:2018} which is not publicly available at the time of writing this manuscript. Table~\ref{tab:corpora} reports the size of each corpus in raw and deduplicated version.

\begin{table}[t]
    \centering
    \small
    \scalebox{0.8}{
    \begin{tabular}{lccccc}
    \toprule
    & \textsc{Raw} & \multicolumn{2}{c}{{\textsc{Deduplication}}} &\multicolumn{2}{c}{{\textsc{CSA Indexing}}}\\
    \cmidrule(lr){2-2}\cmidrule(lr){3-4}\cmidrule(lr){5-6} 
        {\textsc{}} & \textsc{Size} &\textsc{Time}&\textsc{Size }&\textsc{Time}&\textsc{Size}\\
        {\textsc{Corpus}} & \textsc{(GB)} &\textsc{ (Min)}&\textsc{ (GB)}&\textsc{ (Min)}&\textsc{ (GB)}\\
        \midrule
HackerNews & 3.9 &7,147.2& 3.2&34.2& 3.3 \\
BookCorpus& 4.3 &14,301.2& 3.7 & 88.1 &3.6\\
DM Mathematics& 7.8 &7,881.6& 1.7&32.5& 3.7 \\
OpenSubtitles& 13 &19,920.1& 4.9& 58.1&4.8 \\
Guthenberg& 10.9 &23,893.0& 9.7& 139.0 &9.5 \\
Wikipedi& 17 &31,124.4& 14&160.4&13 \\
USPTO& 22.9 &41,866.8& 22 &206.8 &16 \\
OpenWebTexts & 62.8 &115,088.2& 54&885.8 &47 \\
CCNewsv2 & 150 & 292,724.7& 94& 818.3 &80 \\
Pile-CC & 227.1 &416,186.8& 123& 1,965.2 &106\\
Reddit& 420	&617,906.5& 345&4,821.2& 358 \\
         \bottomrule
    \end{tabular}
    }
    \caption{Statistics of corpora, deduplication step, and the index construction. Indexing is done on a single CPU core of a 2.70 GHz Intel Xeon Gold 6150, and requires $2.5\times$ of index size of RAM memory.}
    \label{tab:corpora}
\end{table}
\section{Pipeline and Features of \texttt{Koala}}
\subsection{Data Structure of \texttt{Koala}}\label{sec:index}
Our index construction is inspired by the language models of \citet{Shareghi:2015,Shareghi:2016TACL}, which leverage compressed data structures for building language models on large text corpora. In this subsection we provide a brief overview of the data structures behind \texttt{Koala}. 

\begin{table*}[t]
    \centering
    \small
    \scalebox{0.59}{
    \begin{tabular}{lllllllllllll}
    \toprule
$n$&$n$-grams list&Pile-CC&BookCorpus&CCNewsv2&DM&Guthenberg&HackerNews&OpenSubtitles&OpenWebTexts&USPTO &Wikipedia&Reddit\\\cmidrule(lr){1-1}\cmidrule(lr){2-2}\cmidrule(lr){3-13}
\multirow{6}{*}{\bf 1} &plastic&959364&33845&580607&0&4964&14397&14114&329535&598625&39435&2650049\\
&bags&578401&29213&415672&0&17160&5405&21590&166685&111115&13708&1697726\\
&floating&303836&19752&162095&0&36242&10058&8165&120146&244489&21938&976575\\
&in&355723492&9260245&308475794&3347881&30592137&7135629&7831355&150523086&63002717&54190836&749899124\\
&the&1056004732&34886372&782874590&6519155&107380032&20809865&23296159&428544710&251429575&128120455&2128039302\\
&ocean&575919&30175&273507&0&65172&8467&23233&235331&23909&41516&1125595\\\cdashline{1-13}
\multirow{5}{*}{\bf 2} &plastic bags&39722&843&38094&0&0&588&367&19323&7544&1267&79539\\
&bags floating&77&4&57&0&0&2&2&25&0&5&275\\
&floating in&29619&3326&19189&0&3492&408&1397&12907&2913&1695&101880\\
&in the&91136626&2440752&81218136&52379&7948909&1572721&1925941&37928620&19087529&13710461&175900138\\
&the ocean&284689&18995&139332&0&33275&4066&14749&114465&11596&18558&667336\\\cdashline{1-13}
\multirow{4}{*}{\bf 3} &plastic bags floating&34&0&22&0&0&1&0&12&0&2&110\\
&bags floating in&27&0&34&0&0&0&0&8&0&3&101\\
&floating in the&14481&1621&10734&0&1791&141&725&6594&1760&897&43090\\
&in the ocean&44233&1573&28680&0&2025&1035&2513&21517&1588&2566&163343\\\cdashline{1-13}
\multirow{3}{*}{\bf 4} &plastic bags floating in&16&0&10&0&0&0&0&3&0&2&43\\
&bags floating in the&20&0&29&0&0&0&0&5&0&3&76\\
&floating in the ocean&580&19&413&0&7&10&16&372&24&42&2078\\\cdashline{1-13}
\multirow{2}{*}{\bf 5} &plastic bags floating in the&13&0&8&0&0&0&0&1&0&2&33\\
&bags floating in the ocean&4&0&2&0&0&0&0&1&0&2&9\\\cdashline{1-13}
{\bf 6} &plastic bags floating in the ocean&4&0&1&0&0&0&0&1&0&2&2\\\bottomrule
    \end{tabular}
    }
    \caption{The n-gram hit statistics per corpus for the correct answer (\emph{plastic bags floating in the ocean}) to the query \emph{Which of these situations is an example of pollutants?, choices : [\textbf{plastic bags floating in the ocean}, mallard ducks floating on a lake, cottonwood seeds floating in the air, cirrus clouds floating in the sky]}. This is a sample from the OpenBookQA benchmark.}
    \label{tab:ngramhits}
\end{table*}

A Suffix Array (SA)~\cite{Manber:1993} of a string $\mathcal{T}$ with alphabet $\sigma$ is an array of its sorted suffixes. A cell in a suffix array, denoted by $\text{SA}[i]$, stores a number indicating the staring position of its corresponding suffix in $\mathcal{T}$. Using a suffix array, searching for any sequence $\mathbf{u}$ in $\mathcal{T}$ translates into a binary search to find the range that spans over all substrings that have $\mathbf{u}$ as their prefix, and is $\mathcal{O}(|\mathbf{u}|\log |\mathcal{T}|)$. Constructing SA takes $4$-$8|\mathcal{T}|$ bytes in practice, 
making them impractical to use for large data. 

To support search on large collections, Compressed Suffix Array exploits the compressibility of $\mathcal{T}$ while providing the same functionality of SA in space equal to bzip2 compressed $\mathcal{T}$ in practice. We follow ~\citet{Shareghi:2016TACL} and use the FM-Index~\cite{Ferragina:2008} that utilises the text compressibility vi the Burrows-Wheeler transformation (BWT)~\cite{Burrows:1994} of the text. The BWT is defined as, $\text{BWT}[i] = [\text{SA}[i] - 1\ \text{mod}\ |\mathcal{T}|]$. Searching for a sequence in BWT is done in reverse order and requires $\mathcal{O}(|\mathbf{u}|\log|\sigma|)$. For more details on BWT and reverse searching, refer to ~\citet{nv-csurv07}.

The CSA is at the core of \texttt{Koala}'s index and search backbone. We used the SDSL library \cite{Gog:2014} to implement our corpus indexer. We index each corpus separately. Once a corpus is indexed, its constructed index sits on disk and could be queried through the \texttt{Koala} web interface (introduced shortly). Each query is launched into the indexed collection of corpora and returns the hit counts of the query in the corresponding corpus.
\begin{figure*}
    \centering
    \includegraphics[scale=0.47]{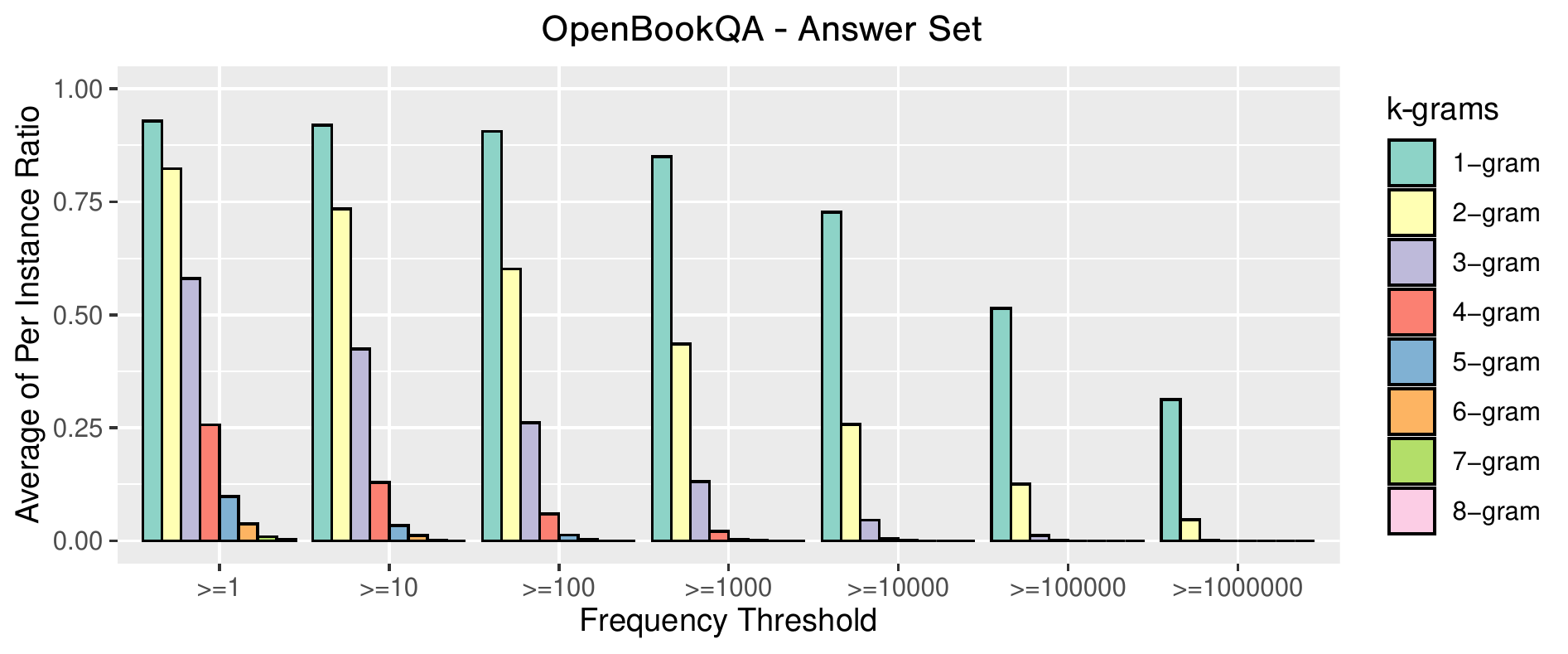}
        \includegraphics[width=6.8cm, keepaspectratio]{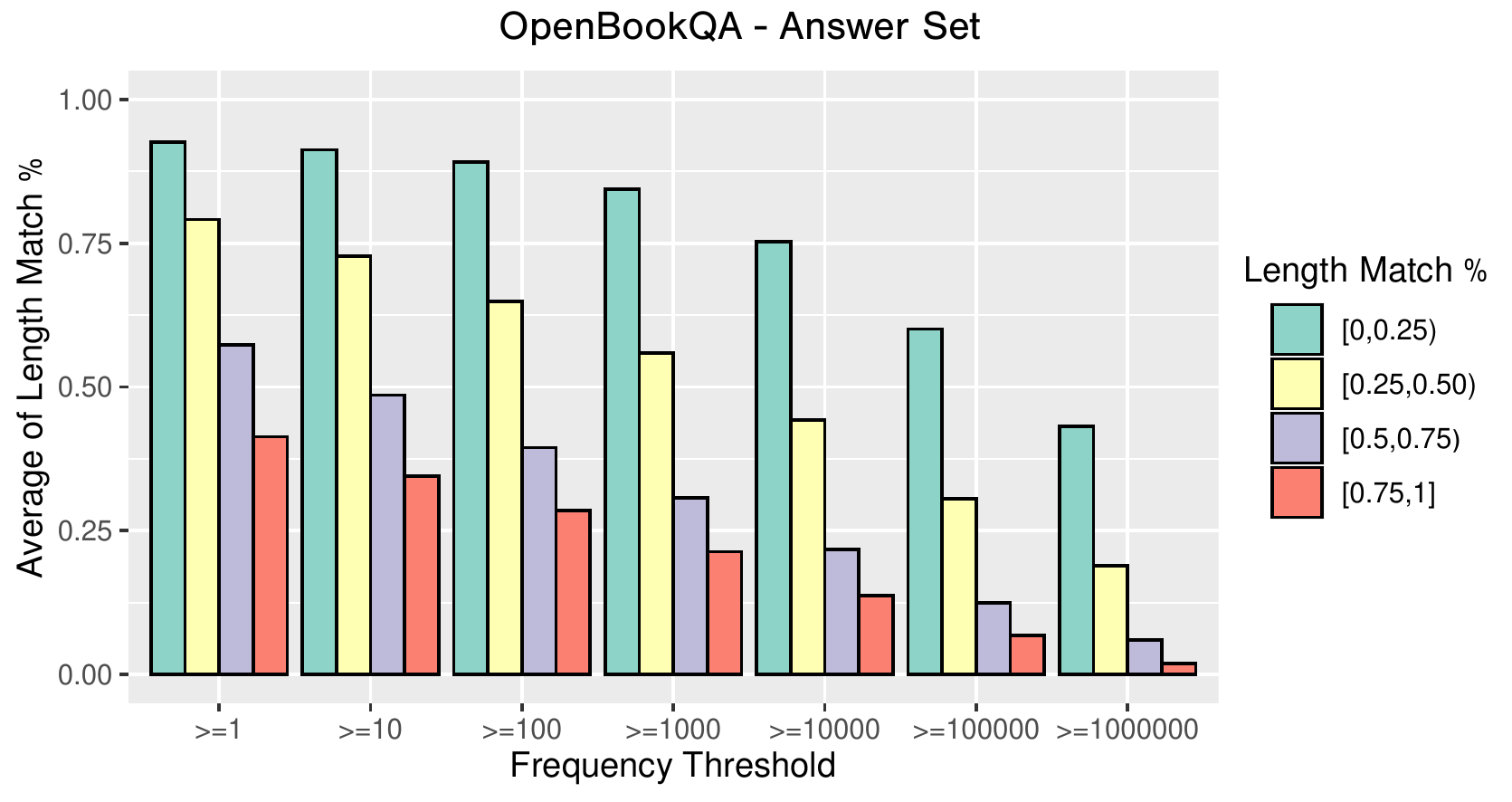}
    \includegraphics[scale=0.47]{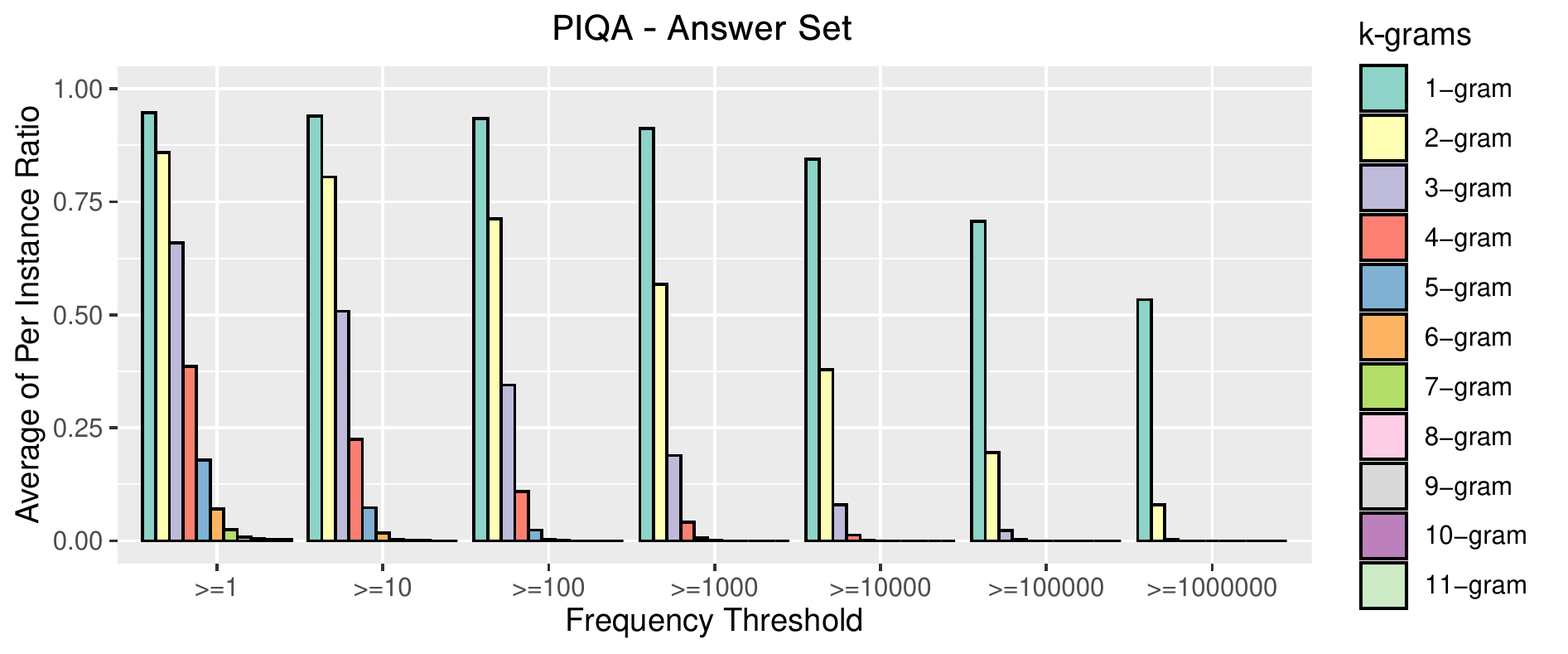}
    \includegraphics[width=6.8cm, keepaspectratio]{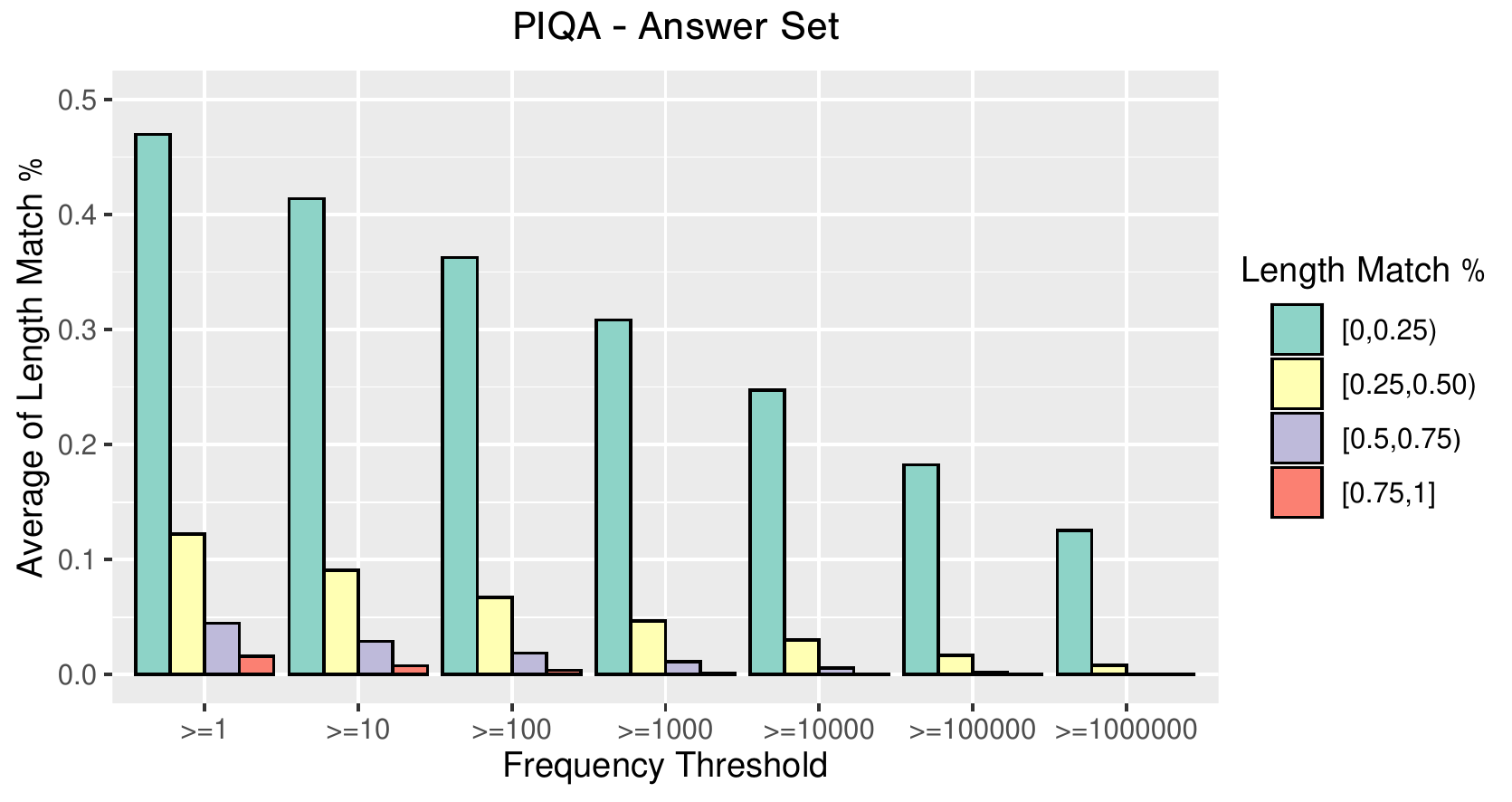}
    \caption{Visualisations of $n$-gram overlap statistics for OpenBookQA and PIQA test sets, Answer side. \textbf{Top:} OpenBookQA Answer Set ; \textbf{Bottom:} PIQA Answer Set. \textbf{Left:} Average of Per Instance K-gram hit ratio (i.e., K-gram hit ratio = 1 means 100\% of k-grams in one instance were a hit); \textbf{Right:} Average of Per Instance K-gram hit length ratio (i.e., K-gram hit length ratio with respect to the instance length = 1 means the k-gram was fully covered, 0.75 means it was 3/4 covered, etc). PIQA test set size is 1838, OpenBookQA test set size is 500.}
    \label{fig:insights}
\end{figure*}
Table~\ref{tab:corpora} reports the time and memory usage for construction of indexes.
\begin{figure*}[t]
         \centering
          \includegraphics[trim={0 0.2cm 0 0},clip, scale=0.57]{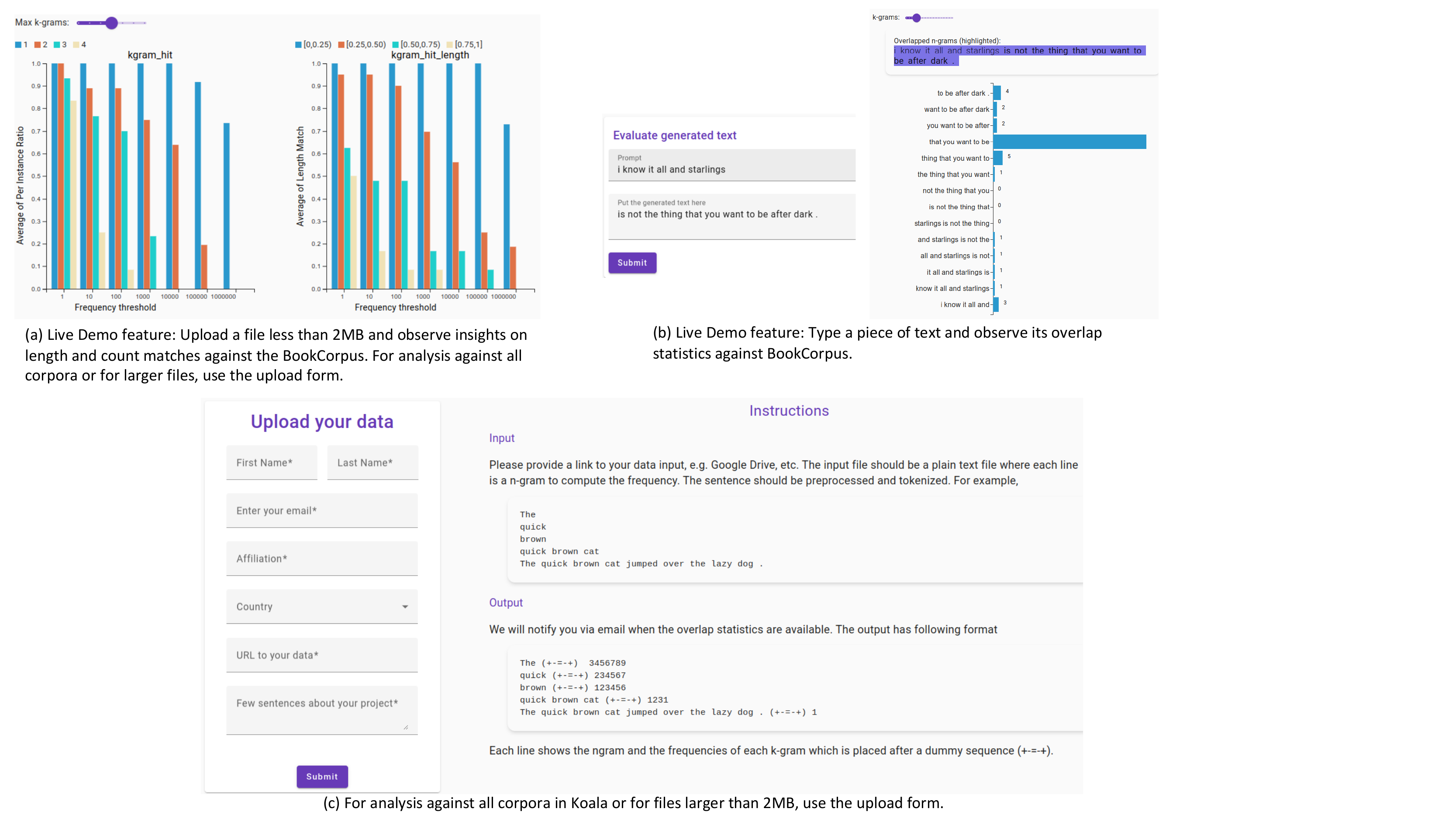}
     \caption{Screenshots from different features of the \texttt{Koala} webpage. For the latest version of the interface, please refer to the website.}
     \label{fig:koala-livedemo}
\end{figure*}

\subsection{$n$-gram Overlap Statistics of \texttt{Koala}} \label{sec:hitratios}
Given a text query, \texttt{Koala} can provide its count statistics in several pretraining corpora by querying the indexes. An example of the raw count output for the phrase \emph{plastic bags floating in the ocean} is shown in Table~\ref{tab:ngramhits}. Meaningful insights can be derived from these raw statistics. Figure~\ref{fig:insights} illustrates two high-level statistics built on top of the $n$-gram counts for two question answering benchmark test sets, PIQA~\citep{Bisk2020} and OpenBookQA~\citep{mihaylov-etal-2018-suit}, highlighting the amount of leakage or overlap that exists between these test sets and the entire pre-training data collection indexed in \texttt{Koala}. We first introduce how these statistics are calculated per instance, noting that Figure~\ref{fig:insights} is reporting them as an average across all instances in each test set. The high-level statistics are defined as follows: 
\begin{description}[noitemsep,leftmargin=2.5mm,topsep=0pt,parsep=0pt,partopsep=0pt]
\item[\emph{Per Instance $k$-gram hit ratio}] measures $\frac{M^{k,t}_x}{N^k_x}$, where $N^k_x$ is the set of all $k$-grams of instance $x$, and ${M^{k,t}_x}$ is the subset of $N^k_x$ containing only the $k$-grams with frequency above the pre-set thresholds $t$ (e.g., $\geq$ 1, $\geq$ 10, $\geq$ 100, $\geq$ 1k, $\geq$ 10k, $\geq$100k, $\geq$1M).
    \item[\emph{Per Instance $k$-gram hit length ratio}] measures
    $\frac{M^{l,t}_x}{N^l_x}$, where $N^l_x$ is the set of all substrings of instance $x$ that fall within the length bin $l$ (e.g., $l=[0.75,100]$ means all substrings whose lengths are 3/4 of the length of $x$ or more), and $M_{x}^{l,t}$ is the subset of $N^l_x$, containing only the substrings with frequency above the pre-set thresholds $t$ (e.g., $\geq$ 1, $\geq$ 10, $\geq$ 100, $\geq$ 1k, $\geq$ 10k, $\geq$100k, $\geq$1M). We considered 4 length bins: [0,0.25), [0.25,0.50), [0.5,0.75), and [0.75,1].
\end{description}
While a deep dive into exploring the dependence between data overlap, model size, and model performance requires a separate work, here we unpack some highlights from the figures:

\paragraph{Highlights from Figure~\ref{fig:insights} (Left Panel):} The top-left panel highlights that for OpenBookQA above 75\% of the unigrams and bigrams of test set occur at least once ($\geq$ 1) in the pretraining data, while this drops to below 50\% with a higher threshold ($\geq$ 1k). We observe that above 25\% of trigrams occur at least $100$ times in the pretraining data. Looking at the bottom-left panel for PIQA, we see a much stronger indication of data overlap. For instance we observe above 55\% of  bigrams occur at least $100$ times in the pre-training data. Comparing the two dataset at the extreme frequency threshold of $\geq$ 1M, we observe that above 50\% of PIQA unigrams occur at least 1M times in the pretraining data, while this is roughly 30\% for OpenBookQA.  

\paragraph{Highlights from Figure~\ref{fig:insights} (Right Panel):} The average answer length in PIQA and OpenBookQA test sets are 101, and 20, respectively. This means that [0.25,0.5) length bin covers sequences of roughly 25-50 tokens for PIQA, while this is roughly 5-10 tokens for OpenBookQA. We now turn to the highlights from the right panel of Figure~\ref{fig:insights}. For OpenBookQA (top-right) we observe from the red bars that above 25\% of test instances (roughly 125 cases out of 500 test instances in OpenBookQA) are almost [75\%,100\%] covered in the pre-training data for at least 100 times ($\geq$ 100). This corresponds to matches of length 15-20 words. Looking at PIQA (Bottom-Right), although the coverage with respect to the full length is not as apparent as OpenBookQA, matches in each corresponding length bin of PIQA are roughly 4$\times$ longer than OpenBookQA. For instance, about 5\% of test instances of PIQA (roughly 90 cases out of 1838 test instances in PIQA) have a matching substring of 25-50 words which occur at least $1k$ times in the pretraining data (see yellow bar for $\geq$ 1000).   

The performance ceiling obtained by GPT-3 and OPT models for these two benchmarks (reported numbers in Appendix A of ~\citet{zhang2022opt}) indicate the largest variant of both models achieve roughly 80\% accuracy for PIQA, and above 57\% accuracy on OpenBookQA. Our highlighted findings suggests a positive correlation between the amount of data overlap we highlighted and the task performance ceiling by the LLMs trained on the same pre-training corpora. As a future direction of analysis, it would be interesting to leverage \texttt{Koala} to analyse the interdependence of the amount of data overlap, model size, and task performance. Our preliminary analyses suggest that the connection between model size and capacity to memorise is not trivial, and varies from task to task.

\subsection{Interface of \texttt{Koala}}
In this section, we give an overview of the interface of \texttt{Koala}. Figure~\ref{fig:koala-livedemo} demonstrates some of \texttt{Koala}'s features. 
%
%
%
In addition to reporting the raw counts, 
\texttt{Koala} provides an interface to upload an $n$-gram file and to visualize different hit ratio statistics (\S\ref{sec:hitratios}). The $n$-gram file is a plain text file where each line is an $n$-gram whose overlap statistics will be computed. Figure~\ref{fig:koala-livedemo}(a) shows the output from this feature. We also provide the interactive version of the ratio plots (e.g.,  Figure~\ref{fig:insights}) for 3  question answering benchmarks: HellaSwag~\citep{zellers-etal-2019-hellaswag}, PIQA~\citep{Bisk2020} and OpenBookQA~\citep{mihaylov-etal-2018-suit}.\footnote{We plan to expand the coverage of benchmarks.}

In the first release, we limit the live demo queries to $n$-gram files below 2MB and report only on the BookCorpus pretraining corpus. For larger files and more comprehensive statistics, we provide a form for users to submit the data and queue the computation. An example of the form is shown in Figure~\ref{fig:koala-livedemo}(c). We plan to extend the live demo to the entire pretraining corpora in the near future.

Another use case of the overlap statistics is to provide a measure of the memorization vs. creativity for generative LLMs, i.e. how much of the generated text overlaps with the pretraining corpora. \texttt{Koala} implements a tool to verify the novelty of an output of generative LLM given a prompt. Figure~\ref{fig:koala-livedemo}(b) shows an example of this feature which provides the count statistics of the $n$-grams in the generated text and highlight the overlap $n$-grams.

\section{Conclusion and Future Work}
We presented \texttt{Koala}, a web-based service powered by a compressed data structure backbone that facilitates efficient search over large collections of texts. \texttt{Koala} is a tool for comprehensive overlap analysis with potential use-cases including but not limited to assessing leakage of test benchmarks,  measuring the degree of memorization in generative LLMs outputs. Additionally, \texttt{Koala} not only provides a public tool for forensic analysis of these phenomena it could also help benchmark designers towards constructing more challenging testbeds for LLMs. We will continue to grow our coverage of the existing pre-training corpora. 
\bibliography{acl2023}
\bibliographystyle{acl_natbib}


\end{document}